# Learning Hierarchical Object Maps Of Non-Stationary Environments With Mobile Robots

Dragomir Anguelov*   Rahul Biswas*   Daphne Koller*
Benson Limketkai*   Sebastian Thrun†

*Computer Science Department
Stanford University
Stanford, CA 94305

†School of Computer Science
Carnegie Mellon University
Pittsburgh, PA 15213

## Abstract

Building models, or maps, of robot environments is a highly active research area; however, most existing techniques construct unstructured maps and assume static environments. In this paper, we present an algorithm for learning object models of non-stationary objects found in office-type environments. Our algorithm exploits the fact that many objects found in office environments look alike (e.g., chairs, recycling bins). It does so through a two-level hierarchical representation, which links individual objects with generic shape templates of object classes. We derive an approximate EM algorithm for learning shape parameters at both levels of the hierarchy, using local occupancy grid maps for representing shape. Additionally, we develop a Bayesian model selection algorithm that enables the robot to estimate the total number of objects and object templates in the environment. Experimental results using a real robot equipped with a laser range finder indicate that our approach performs well at learning object-based maps of simple office environments. The approach outperforms a previously developed non-hierarchical algorithm that models objects but lacks class templates.

## 1 Introduction

Building environmental maps with mobile robots is a key prerequisite of truly autonomous robots [19]. State-of-the-art algorithms focus predominantly on building maps in *static* environments [20]. Common map representations range from lists of landmarks [3, 9, 21], fine-grained grids of numerical occupancy values [6, 15], collections of point obstacles [11], or sets of polygons [12]. These representations are appropriate for mobile robot navigation in static environments.

Real environments, however, consist of objects. For example, office environments possess chairs, doors, recycling bins, etc. Many of these objects are non-stationary, that is, their locations may change over time. This observation motivates research on a new generation of mapping algorithms, which represent environments as collections of objects. At a minimum, such object models would enable a robot to track changes in the environment. For example, a cleaning robot entering an office at night might realize that a recycling bin has moved from one location to another. It might do so without the need to learn a model of this recycling bin *from scratch*, as would be necessary with existing robot mapping techniques [20].

Object representations offer a second, important advantage, which is due to the fact that many office environments possess large collections of objects of the same type. For example, most office chairs are instances of the same generic chair and therefore look alike, as do most doors, recycling bins, and so on. As these examples suggest, attributes of objects are shared by entire classes of objects, and understanding the nature of object classes is of significant interest to mobile robotics. In particular, algorithms that learn properties of object classes would be able to transfer learned parameters (e.g., appearance, motion parameters) from one object to another in the same class. Such ability to generalize would have a profound impact on the accuracy of object models, and the speed at which such models can be acquired. If, for example, a cleaning robot enters a room it has never visited before, it might realize that a specific object in the room possesses the same visual appearance of other objects seen in other rooms (e.g., chairs). The robot would then be able to acquire a map of this object much faster. It would also enable the robot to predict properties of this newly seen object, such as the fact that a chair is non-stationary—without ever seeing this specific object move.

In previous work, we developed an algorithm that has successfully been demonstrated to learn shape models of non-stationary objects [2]. This approach works by comparing occupancy grid maps acquired at different points in time. A straightforward segmentation algorithm was developed that extracts object footprints from occupancy grid maps. It uses these footprints to learn shape models of objects in the environment, represented by occupancy grid maps. This algorithm is related to work on learning generative object models in computer vision and medical imaging. Frey and Jojic [7] describe an unsupervised approach which infers a set of object templates and their transformations from a set of camera images. Leventon et al. [10] describe an alternative shape representation based on geodesic active contours. They show how to learn ob-ject shape priors using the representation and how to use the object priors for tissue segmentation in tomography scans.

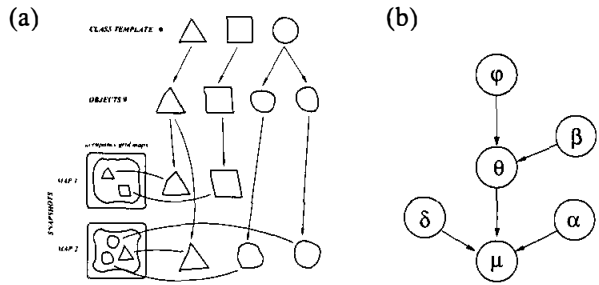

Figure 1: (a) Generative hierarchical model of environments with non-stationary objects. (b) Representation as a graphical model.

This paper goes one step further by proposing an algorithm that identifies classes of objects, in addition to learning plain object models. In particular, our approach learns shape models of individual object classes, from multiple occurrences of objects of the same type. By learning shape models of object types—in addition to shape models of individual objects—our approach is able to generalize across different object models that belong to the same object class. This approach follows the hierarchical Bayesian framework (see [1, 8, 13]). We show that our approach leads to significantly more accurate models in real-world environments with multiple objects of the same type.

The specific learning algorithm proposed here is an instance of the popular EM algorithm [14]. We develop a closed-form solution for learning at both levels of the hierarchy, which simultaneously identifies object models and shape templates for entire object classes. On top of this, we propose a Bayesian procedure for determining the appropriate number of object models, and object class models.

We tested our algorithm on data gathered by a physical robot, which was equipped with a laser range finder. Our results suggest that our approach succeeds in learning accurate shape and class models. A systematic comparison with our previous, non-hierarchical approach [2] illustrates that the use of class models yields significantly better results, both in terms of predictive power (as measured by the log-likelihood over testing data) and in terms of convergence properties (measured by the number of times each algorithm is trapped in a local maximum of poor quality).

## 2 The Generative Hierarchical Model

We begin with a description of the hierarchical model. The object level generalizes the approach of [2] to maps with continuous occupancy. The central innovation is the introduction of a template level.

### 2.1 The Object Hierarchy

Our object hierarchy (Figure 1a) is composed of two levels, the *object template level* at the top, and the *physical object level* at the bottom. The object template level consists of a set of $M$ shape templates, denoted $\varphi = \varphi_1, \ldots, \varphi_M$. Each template $\varphi_m$ is represented by an occupancy grid map [6, 15, 20], that is, an array of values in $[0, 1]$ that represent the occupancy of a grid cell.

The object level contains shape models of concrete objects in the world, denoted: $\theta = \theta_1, \ldots, \theta_N$, where $N$ is the total number of objects (with $N \geq M$). Each object model $\theta_n$ is represented by an occupancy grid map, just like at the template level. The key difference between object models $\theta_n$ and templates $\varphi_m$ is that each $\theta_n$ corresponds to exactly one object in the world, whereas a template $\varphi_m$ may correspond to more than one object. If, for example, all non-stationary objects were to look alike, $\theta$ would contain multiple models (one for each object), whereas $\varphi$ would contain only a single shape template.

To learn a hierarchy, we assume that the robot maps its environments at $T$ different points in time, between which the configuration of the environment may have changed. Each map is obtained from laser sensor readings and is represented as a (static) occupancy grid map. The sequence of maps is denoted $\mu = \mu_1, \ldots, \mu_T$.

Objects may or may not be present at any time $t$, and they may be located anywhere in the free space of the environment. The number of object snapshots present in the map $\mu_t$ is denoted $K_t$. The set of object snapshots extracted from the map $\mu_t$ are denoted $\mu_t = \mu_{1,t}, \ldots, \mu_{K_t,t}$.

Each object snapshot $\mu_{k,t}$ is—once again—represented by an occupancy grid map, constructed from robot sensor measurements [6, 15, 20]. The exact routines for extraction of object snapshots from maps are described in [2] and will be reviewed briefly below.

Finally, we notice that objects may be observed in any orientation. Since aligning object snapshots with objects in the model is an important step in the learning procedure, we will make the alignment parameters explicit. In particular, we will use $\delta_{k,t}$ to denote the alignment of snapshot $\mu_{k,t}$ relative to the generative model. In our implementation, each $\delta_{k,t}$ consists of two translational and one rotational parameters.

### 2.2 Probabilistic Model

To devise a sound algorithm for inferring an object hierarchy from data, we have to specify probabilistic models of how snapshots are generated from objects and how objects are generated from object templates. An graphical overview for our probabilistic model is shown in Figure 1b.

Let $\theta_n$ be a concrete object, and $\mu_{k,t}$ be a single snapshot of this object. Recall that each grid cell $\theta_n[j]$ in $\theta_n$ is a real number in the interval $[0, 1]$. We interpret each occupancy value as a probability that a robot sensor would detect an occupied grid cell. However, when mapping an environment, the robot typically takes multiple scans of the same object, each resulting in a binomial outcome. By aggregating the individual binary variables into a single aggregate real value, we can approximate this fairly cumbersome model into a much cleaner Gaussian distribution of a single real-valued observation. Thus, the probability of observing a concrete snapshot $\mu_{k,t}$ of object $\theta_n$ is given by

$$p(\mu_{k,t} \mid \theta_n, \delta_{k,t}) \propto e^{-\frac{1}{2\rho^2} \sum_j (f(\mu_{k,t}, \delta_{k,t})[j] - \theta_n[j])^2} \quad (1)$$

The function $f(\mu_{k,t}, \delta_{k,t})$ denotes the aligned snapshot $\mu_{k,t}$, and $f(\mu_{k,t}, \delta_{k,t})[j]$ denotes its $j$-th grid cell. The rotation and translation parameters of the alignment are specified by $\delta_{k,t}$. The parameter $\rho^2$ is the variance of the noise.

It is useful to make explicit the correspondence between objects $\theta_n$ and object snapshots $\mu_{k,t}$, by introducing *correspondence variables* $\alpha = \alpha_1, \alpha_2, \ldots, \alpha_T$. Since each $\mu_t$ is an entire set of snapshots, each $\alpha_t$ is in fact a function: $\alpha_t : \{1, \ldots, K_t\} \longrightarrow \{1, \ldots, N\}$.

A similar model governs the relationship between templates and individual objects. Let $\theta_n$ be a concrete object generated according to object template $\varphi_m$, for some $n$ and $m$. The probability that a grid cell $\theta_n[j]$ takes on a value $r \in [0, 1]$ is a function of the corresponding grid cell $\varphi_m[j]$. We assume that the probability of a grid cell value $\theta_n[j]$ is normally distributed with variance $\sigma^2$:

$$p(\theta_n[j] \mid \varphi_m[j]) = \frac{1}{\sqrt{2\pi}\sigma} e^{-\frac{1}{2\sigma^2}(\theta_n[j] - \varphi_m[j])^2} \quad (2)$$

Equation (2) defines a probabilistic model for individual grid cells, which is easily extended to entire maps:

$$\begin{aligned} p(\theta_n \mid \varphi_m) &= \prod_j p(\theta_n[j] \mid \varphi_m[j]) \\ &\propto e^{-\frac{1}{2\sigma^2} \sum_j (\theta_n[j] - \varphi_m[j])^2} \end{aligned} \quad (3)$$

Again, we introduce explicit variables for the correspondence between objects $\theta_n$ and templates $\varphi_m$: $\beta = \beta_1, \ldots, \beta_N$ with $\beta_n \in \{1, \ldots, M\}$. The statement $\beta_n = m$ means that object $\theta_n$ is an instantiation of the template $\varphi_m$. The correspondences $\beta$ are *unknown* in the hierarchical learning problem, which is a key complicating factor in our attempt to learn hierarchical object models.

There is an important distinction between the correspondence variables $\alpha$'s and $\beta$'s, arising from the fact that each object $\theta_n$ can only be observed once when acquiring a local map $\mu_t$. This induces a *mutual exclusivity constraint* on the set of valid correspondences at the object level: If $k \neq k'$, then $\alpha_t(k) \neq \alpha_t(k')$. Thus, we see that the physical objects, modeled in the object level, can only be observed at most once in any given map, whereas the class level object templates might be instantiated more than once. For example, an object at the class level might be a prototypical chair, which might be mapped to multiple concrete chairs at the object level—and usually multiple observations over time of any of those concrete chairs at the snapshot level.

## 3 Hierarchical EM

Our goal in this paper is to learn the model $\Psi = \langle \theta, \varphi, \delta \rangle$ given the data $\mu$ using EM [5]. Unlike many EM implementations, however, we do not simply want to maximize the probability of the data given the model. We also want to take into consideration the probabilistic relationships between the two levels of the hierarchy. Thus, we want to maximize the joint probability over the data $\mu$ and the model $\Psi$:

$$\underset{\Psi}{\mathrm{argmax}}\, p(\Psi, \mu) = \underset{\theta, \varphi, \delta}{\mathrm{argmax}}\, p(\theta, \varphi, \delta, \mu) \quad (4)$$

Note that we treat the latent alignment parameters $\delta$ as model parameters, which we maximize during learning.

EM is an iterative procedure that can be used to maximize a likelihood function. Starting with some initial model, EM generates a sequence of models of non-decreasing likelihood:

$$\langle \theta^{[1]}, \varphi^{[1]}, \delta^{[1]} \rangle, \langle \theta^{[2]}, \varphi^{[2]}, \delta^{[2]} \rangle, \ldots \quad (5)$$

Let $\Psi^{[i]} = \langle \theta^{[i]}, \varphi^{[i]}, \delta^{[i]} \rangle$ be the $i$-th such model. Our desire is to find an $(i+1)$st model $\Psi^{[i+1]}$ for which

$$p(\Psi^{[i+1]}, \mu) \geq p(\Psi^{[i]}, \mu) \quad (6)$$

As is common in the EM literature [14], this goal is achieved by maximizing the expected log likelihood

$$\Psi^{[i+1]} = \underset{\Psi}{\mathrm{argmax}}\, E_{\alpha, \beta} \left[ \log p(\alpha, \beta, \Psi, \mu) \, \Big| \, \Psi^{[i]}, \mu \right] \quad (7)$$

Here $E_{\alpha, \beta}$ is the mathematical expectation over the latent correspondence variables $\alpha$ and $\beta$, relative to the distribution $p(\alpha, \beta \mid \Psi^{[i]}, \mu)$.

The probability inside the logarithm in (7) factors into two terms, one for each level of the hierarchy (multiplied by a constant):

$$p(\alpha, \beta, \Psi, \mu) = p(\alpha, \beta, \varphi, \theta, \delta, \mu) \quad (8)$$

Exploiting the independencies shown in Figure 1b, and the uniform priors over $\phi$, $\alpha$, and $\beta$, we obtain:

$$\begin{aligned} &= p(\varphi)\, p(\beta)\, p(\theta \mid \beta, \varphi)\, p(\alpha)\, p(\delta)\, p(\mu \mid \delta, \alpha, \theta) \\ &\propto p(\theta \mid \beta, \varphi)\, p(\mu \mid \delta, \alpha, \theta) \end{aligned} \quad (9)$$

The probability $\log p(\theta \mid \beta, \varphi)$ of the objects $\theta$ given the object templates $\varphi$ and the correspondences $\beta$ is essentially defined via (3). Here we recast it using a notation that makes the conditioning on $\beta$ explicit:

$$p(\theta \mid \beta, \varphi) \quad (10)$$

$$\propto \prod_{n=1}^{N} e^{-\frac{1}{2\sigma^2} \sum_{m=1}^{M} I(\beta_n = m) \sum_j (\theta_n[j] - \varphi_m[j])^2}$$

where $I(\ )$ is an indicator function which is 1 if its argument is true, and 0 otherwise. Similarly, the probability $p(\mu \mid \alpha, \theta, \delta)$ is based on (1) and conveniently written as:

$$p(\mu \mid \alpha, \theta, \delta) \propto \quad (11)$$

$$\prod_{t=1}^{T} \prod_{k=1}^{K_t} e^{-\frac{1}{2\rho^2} \sum_{n=1}^{N} I(\alpha_t(k) = n) \sum_j (f(\mu_{k,t}, \delta_{k,t})[j] - \theta_n[j])^2}$$

Substituting the product (9) with (10) and (11) into the expected log likelihood (7) gives us:

$$\Psi^{[i+1]} = \underset{\varphi,\theta,\delta}{\operatorname{argmax}}$$

$$-\sum_{n=1}^{N}\left\{\sum_{m=1}^{M}\frac{p(\beta_n=m \mid \Psi^{[i]},\mu)}{\sigma^2}\sum_j(\theta_n[j]-\varphi_m[j])^2 \right. \quad (12)$$

$$\left. +\sum_{t=1}^{T}\sum_{k=1}^{K_t}\frac{p(\alpha_t(k)=n \mid \Psi^{[i]},\mu)}{\rho^2}\sum_j(f(\mu_{k,t},\delta_{k,t})[j]-\theta_n[j])^2\right\}$$

In deriving this expression, we exploit the linearity of the expectation, which allows us to replace the indicator variables through probabilities (expectations). It is easy to see that the expected log-likelihood in (12) consists of two main terms. The first enforces consistency between the template and the object level, and the second between the object and the data level.

## 4 The Implementation of the EM Algorithm

The standard implementation of EM requires the M-step to find the parameter assignment $\langle \Psi^{[i+1]}\rangle$ which maximizes (12). A variation of EM called *Generalized EM* [5] requires only that the M-step finds an assignment for $\langle \Psi^{[i+1]}\rangle$ which *increases*, but does not necessarily maximize, the expected log-likelihood in (7). Generalized EM has the convergence guarantees of EM, while possibly taking more iterations to converge. In our case, it allows us to avoid solving a complex joint optimization problem for the model parameters in the M-step.

Our Generalized EM starts with a random model and random alignment parameters. It then alternates an E-step, in which the expectations of the correspondences are calculated given the $i$-th model and alignment, and two M-steps, one that generates a new hierarchical model $\langle \theta^{[i+1]},\varphi^{[i+1]}\rangle$ and one for finding new alignment parameters $\delta^{[i+1]}$. Each of the two M-steps does not decrease the expected log-likelihood. If at least one of them increases it, we have a guarantee of improving the original likelihood in (4). If both result in no change in the expected log-likelihood, then our algorithm has converged. As our objective function is not convex (because of the non-linear projection $f$), it is possible in principle that our implementation of Generalized EM converges to a ridge point, where optimizing only for $\langle \theta^{[i+1]},\varphi^{[i+1]}\rangle$ or for $\delta^{[i+1]}$ does not increase the log-likelihood, while jointly optimizing all the above parameters does. Nevertheless, our experiments show that the algorithm converges rapidly to a good result.

### 4.1 E-Step

In our case, the E-step can easily be implemented exactly:

$$b_{n,m}^{[i]} = p(\beta_n=m \mid \theta^{[i]},\varphi^{[i]})$$

$$= \frac{p(\theta^{[i]} \mid \beta_n=m,\varphi^{[i]})\,p(\beta_n=m \mid,\varphi^{[i]})}{\sum_{m'=1}^{M} p(\theta^{[i]} \mid \beta_n=m',\varphi^{[i]})\,p(\beta_n=m' \mid,\varphi^{[i]})}$$

$$= \frac{p(\theta_n^{[i]} \mid \beta_n=m,\varphi_m^{[i]})}{\sum_{m'=1}^{M} p(\theta_n^{[i]} \mid \beta_n=m',\varphi_m^{[i]})}$$

$$= \frac{e^{-\frac{1}{2\sigma^2}\sum_j(\theta_n^{[i]}[j]-\varphi_m^{[i]}[j])^2}}{\sum_{m'=1}^{M} e^{-\frac{1}{2\sigma^2}\sum_j(\theta_n^{[i]}[j]-\varphi_{m'}^{[i]}[j])^2}} \quad (13)$$

and, similarly,

$$a_{k,t,n}^{[i]} = p(\alpha_t(k)=n \mid \mu^{[i]},\theta^{[i]},\delta_{k,t}) = \quad (14)$$

$$\frac{\sum_{\alpha_t} I(\alpha_t(k)=n)\,e^{-\frac{1}{2\sigma^2}\sum_j \sum_{k'}(f(\mu_{k',t}^{[i]}[j],\delta_{k',t})-\theta_{\alpha_t(k')}^{[i]}[j])^2}}{\sum_{\alpha_t} e^{-\frac{1}{2\sigma^2}\sum_j \sum_{k'}(f(\mu_{k',t}^{[i]}[j],\delta_{k',t})-\theta_{\alpha_t(k')}^{[i]}[j])^2}}$$

The summation over $\alpha_k$ in calculating the expectations $a_{k,t,n}^{[i]}$ is necessary because of the mutual exclusion constraint described above, namely that no object can be seen twice in the same map. The summation is exponential in the number of observed objects $K_t$—however, $K_t$ is rarely larger than 10. If summing over $\alpha_t$ (because of its exponential domain) becomes too costly, efficient (and provably polynomial) sampling schemes can be applied for approximating the desired expectations [4, 16].

### 4.2 Model M-Step

Our M-step first generates a new hierarchical model $\theta, \varphi$ by maximizing (12) under fixed expectations $b_{n,m}^{[i]}$ and $a_{k,t,n}^{[i]}$ and fixed alignment parameters $\delta$. It is an appealing property of our model that this part of the M-step can be executed efficiently in closed form.

Our first observation is that the expression in (12) decomposes into a set of decoupled optimization problems over individual pixels, that can be optimized for each pixel $j$ individually:

$$\langle \theta_n^{[i+1]}[j], \varphi_m^{[i+1]}[j]\rangle = \underset{\theta_n[j]\varphi_m[j]}{\operatorname{argmin}}$$

$$\sum_{n=1}^{N}\sum_{m=1}^{M}\frac{b_{n,m}^{[i]}}{\sigma^2}(\theta_n[j]-\varphi_m[j])^2 \quad (15)$$

$$+ \sum_{n=1}^{N}\sum_{t=1}^{T}\sum_{k=1}^{K_t}\frac{a_{k,t,n}^{[i]}}{\rho^2}(f(\mu_{k,t},\delta_{k,t}^{[i]})[j]-\theta_n[j])^2$$

We then observe that (15) is a quadratic optimization problem, which therefore possesses a convenient closed-form solution [18]. In particular, we can reformulate (15) as a standard least-squares optimization problem:

$$\underset{x[j]}{\operatorname{argmin}}\ (A \cdot x[j] - w[j])^2 \quad (16)$$

where $x[j] = (\theta[j],\varphi[j])$ is a vector comprising the $j$-th cell value of all models at both levels. The constraint matrix $A$ has the form

$$A = \begin{pmatrix} \sigma^{-1}B_{n,m}:\theta & -\sigma^{-1}B_{n,m}:\varphi \\ \rho^{-1}A_{k,t,n} & 0 \end{pmatrix} \quad (17)$$

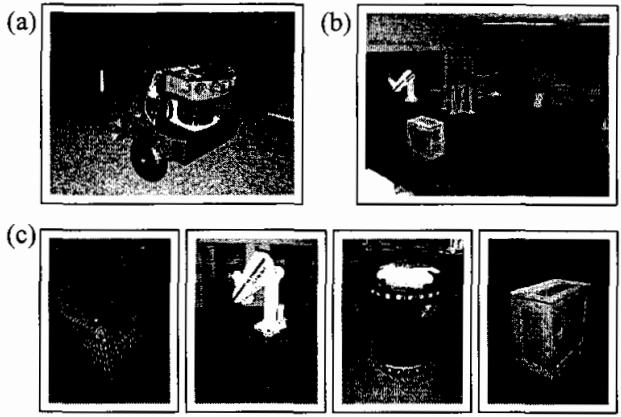

Figure 2: (a) The Pioneer robot used to collect laser range data. (b) The robotics lab where the second data set was collected. (c) Actual images of dynamic objects used in the second data set.

where $B_{n,m}{:}\theta$, $B_{n,m}{:}\varphi$ and $A_{k,t,n}$ are submatrices generated from the expectations calculated in the E-step. Generating such matrices from a quadratic optimization problem such as (15) is straightforward, and the details are omitted here due to space limitations. The vector $w[j]$ is of the form

$$w[j] = \begin{pmatrix} 0 & \rho^{-1} A_{k,t,n} \mu'[j]^T \end{pmatrix} \quad (18)$$

where $\mu'[j]$ is a vector constructed from the aligned $j$-th map cell values of the snapshots $\mu$. The solution to (15) is

$$x[j] = (A^T A)^{-1} A^T w[j] \quad (19)$$

Thus, the new model $\theta_n^{[i+1]}, \varphi_m^{[i+1]}$ is the result of a sequence of simple matrix operations, one for each pixel $j$.

### 4.3 Alignment M-Step

A final step of our M-step involves the optimization of the alignment parameters $\delta$. Those are obtained by maximizing the relevant parts of the expected log likelihood (12). Of significance is the fact that the alignment variables depend only on the object level $\theta$, and not on the template level $\varphi$. This leads to a powerful decomposition by which each $\delta_{k,t}$ can be calculated separately, by minimizing:

$$\delta_{k,t}^{[i+1]} = \underset{\delta_{k,t}}{\operatorname{argmin}} \quad (20)$$
$$\sum_{n=1}^{N} e_{k,t,n}^{[i]} \sum_j (f(\mu_{k,t}, \delta_{k,t})[j] - \theta_n^{[i+1]}[j])^2$$

We represent the $\delta$ for each snapshot as a discrete set of possible transformation values and pick the value of $\delta$ which minimizes the above term. We can use gradient descent to additionally refine the estimate.

### 4.4 Improving Global Convergence

Our approach inherits from EM the property that it is a hill climbing algorithm, subject to local maxima. In our experiments, we found that a straightforward implementation of EM frequently led to suboptimal maps. Our algorithm therefore employs *deterministic annealing* [17] to smooth the likelihood function and improve convergence. In our case, we anneal by varying the noise variance $\sigma$ and $\rho$ in the sensor noise model. Larger variances induce a smoother likelihood function, but ultimately result in fuzzier shape models. Smaller variances lead to crisper maps, but at the expense of an increased number of sub-optimal local maxima. Consequently, our approach anneals the covariance slowly towards the desired values of $\sigma$ and $\rho$, using large values for $\sigma_0$ and $\rho_0$ that are gradually annealed down with an annealing factor $\gamma < 1$:

$$\sigma^{[i]} = \sigma + \gamma^i \sigma_0 \quad (21)$$
$$\rho^{[i]} = \rho + \gamma^i \rho_0 \quad (22)$$

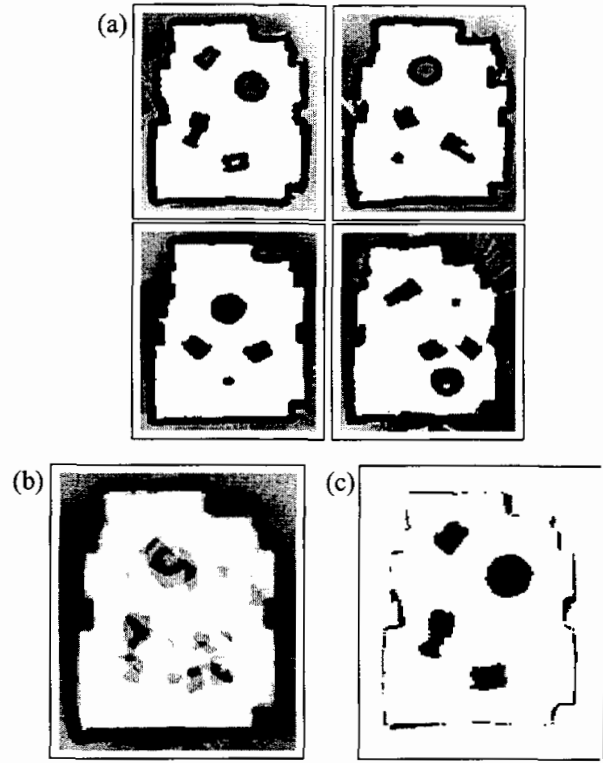

Figure 3: (a) Four maps from the *Study Room* dataset. Each map contains the same set of non-stationary objects. (b) Overlay of optimally aligned maps. (c) A particular difference map *before* low-pass filtering.

The values $\sigma^{[i]}$ and $\rho^{[i]}$ are used in the $i$-th iteration of EM.

### 4.5 Determining the Number of Objects

A final and important component of our mapping algorithm determines the number of class templates $M$ and the number of objects $N$. So far, we have silently assumed that both $M$ and $N$ are given. In practice, both values are unknown and have to be estimated from the data.

The number of objects is bounded below by the number of objects seen in each individual map, and above by the

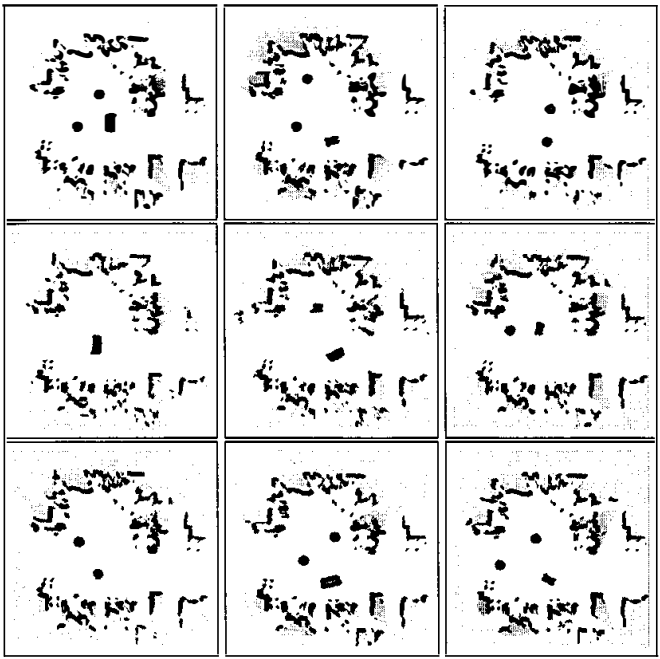

Figure 4: Nine maps from the *Robotics Lab* dataset. The number of objects present varies.

sum of all objects ever seen:

$$\max_t K_t \le N \le \sum_t K_t \quad (23)$$

The number of class templates $M$ is upper-bounded by the number of objects $N$.

Our approach applies a Bayesian prior for selecting the right $N$ and $M$, effectively transforming the learning problem into a *maximum a posteriori (MAP)* estimation problem. At both levels, we use an exponential prior, which in log-form penalizes the log-likelihood in proportion to the number of objects $N$ and object templates $M$:

$$E_{\alpha,\beta}[\log p(\mu, \alpha, \beta \mid \theta, \varphi) \mid \mu, \theta, \varphi] - c_\theta N - c_\varphi M \quad (24)$$

with appropriate constant penalties $c_\theta$ and $c_\varphi$. Hence, our approach applies EM for plausible values of $N$ and $M$. It finally selects those values for $N$ and $M$ that maximize (24), through a separate EM optimization for each value of $N$ and $M$. At first glance this exhaustive search procedure might appear computationally wasteful, but in practice $N$ is usually small (and $M$ is even smaller), so that the optimal values can be found quickly.

## 5　Experimental Results

Our algorithm was evaluated extensively using data collected by a Pioneer robot equipped with a laser range finder. As in [2], maps were acquired in two different office environments: the *Study Room* and the *Robotics Lab*. Figure 2 shows the robot, and some of the non-stationary objects encountered by the robot. Figures 3a and 4 show four and nine example maps extracted in these environments, respectively. Each static map of the *Study Room* always contained the same objects, while in the maps of the *Robotics Lab* all the objects were not necessarily present.

The maps were generated by the concurrent mapping and localization algorithm described in [20]. The individual object snapshots were extracted from regular occupancy grid maps using *map differencing*, a technique closely related to image differencing, which is commonly used in the field of computer vision. In particular, our approach identifies occupied grid cells which, at other points in time, were free. Such cells are candidates of snapshots of moving objects. A subsequent low-pass filter removes artifacts that occur along the boundary of occupied and free space. Finally, a flood-filling technique identifies distinct object snapshots [22]. Empirically, our approach found all non-stationary objects with high reliability as long as they are spaced apart by at least one grid cell (5 cm). Figure 3b shows a typical overlay of the individual maps, and Figure 3c provides examples of object snapshots extracted from those maps. Clearly, more sophisticated methods are needed if objects can touch each other.

In a first series of experiments, we trained our hierarchical model from data collected in the two robot environments. Figure 5a shows an example run of EM for the *Study Room* environment, using the correct number of $N = 4$ objects and $M = 3$ shape templates. As is easily seen, the final object models are highly accurate—in fact, they are more accurate than the individual object snapshots used for their construction. In a series of 20 experiments using different starting points, we found that the hierarchical model converges in all cases to a model of equal quality, whose result is visually indistinguishable from the one presented here. We also tested the ability of our algorithm to correctly associate object snapshots with their object models, and object models with their templates. Figure 5b shows a graph of the probabilities for $\alpha$ and $\beta$ correspondence variables, over iterations of the EM algorithm. As we can see, the model rapidly converges to a definite correspondence, which is the right one. These results are typical for other correspondences.

We then compared our approach with the non-hierarchical technique described in [2]. The purpose of these experiments was to quantify the relative advantage of our hierarchical object model over a shallow model that does not allow for cross-object transfer. We noticed several deficiencies of the non-hierarchical model. The resulting object models were systematically inferior to those generated using our hierarchical approach. Figure 5c shows two examples of results, obtained with different initial random seeds. While the first of these results looks visually adequate, the second does not — it contains an incorrect collection of objects (three circles, one box). Unfortunately, in 11 out of 20 runs, the flat approach converged to such a suboptimal solution.

Moreover, even the visually accurate non-hierarchical models turn out to be inferior. Figure 7 plots log-likelihood

(i) Hierarchical learning, template models

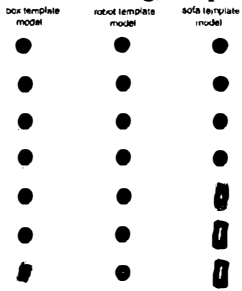

(ii) Hierarchical learning, object models

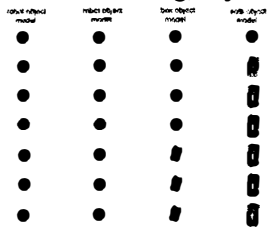

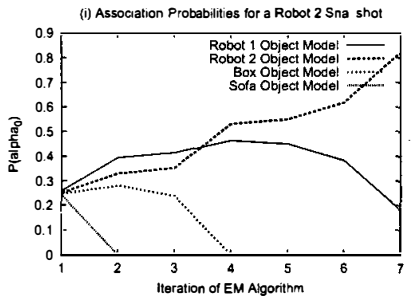

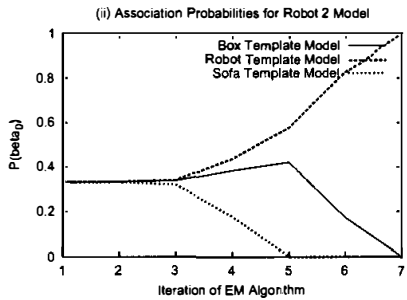

(i) Flat learning, good result

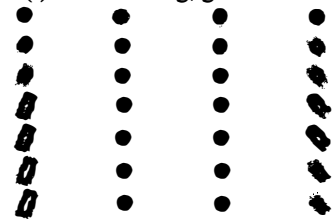

(ii) Flat learning, poor local maximum

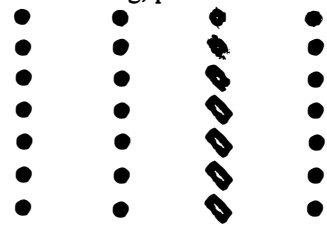

Figure 5: Convergence over seven iterations of EM: (a) Results of the hierarchical learning: (i) templates models, and (ii) object models. (b) Correspondence probabilities (i) between a robot snapshot and different object models (ii) between a robot object model and different template models. (c) Results of the flat model: (i) successful convergence, (ii) unsuccessful convergence to a poor model; 11 out of 20 runs converged poorly.

results on training and testing data for each environment. We perform leave-one-out cross-validation, where we train $T$ different models by leaving one of the $T$ maps in the dataset out. For each model we compute the log-likelihood of both the training and test data and plot these log-likelihood values averaged over the $T$ models. Even in the case when the non-hierarchical approach produces visually adequate results, their actual accuracy lags significantly behind that of the models generated by our hierarchical algorithm. We attribute this difference to the fact that the non-hierarchical approach lacks cross-object generalization.

Finally, we evaluated our approach to model selection, estimating how well our approach can determine both the right number of objects and class templates. Throughout all of our experiments we used the penalty term $35N + 15M$. For both data sets, we found that the log posterior shows a clear peak at the correct values. The results for the *Robotics Lab* are shown in Figure 6, with the correct values shown in bold face. Note that the algorithm converged to the correct value of $N = 4$, although none of the training maps possessed all the 4 objects. The number had to be estimated exclusively based on the fact that, over time, the robot faced 4 different objects with 3 different shapes.

In summary, our experiments indicate that our algorithm learns highly accurate shape models at both levels of the hierarchy, and it consistently identifies the 'right' number of objects and object templates. In comparison with the flat approach described in [2], it yields significantly more accurate object models and also converges more frequently

|       | $N=3$  | $N=4$   | $N=5$  | $N=6$  | $N=7$  |
|-------|--------|---------|--------|--------|--------|
| $M=1$ | −636.2 | −603.1  | −635.5 | −663.9 | −702.4 |
| $M=2$ | −603.2 | −551.5  | −568.7 | −580.7 | −615.2 |
| **M=3** | −612.5 | **−535.7** | −567.5 | −586.9 | −623.2 |
| $M=4$ |        | −550.7  | −587.4 | −567.6 | −618.5 |
| $M=5$ |        |         | −595.0 | −585.4 | −599.2 |
| $M=6$ |        |         |        | −621.3 | −643.1 |
| $M=7$ |        |         |        |        | −608.9 |

Figure 6: Model selection results for the Robotics Lab

to an accurate solution.

## 6 Conclusion

We have presented an algorithm for learning a hierarchy of object models of non-stationary objects with mobile robots. Our approach is based on a generative model which assumes that objects are instantiations of object templates, and are observed by mobile robots when acquiring maps of its environments. An approximate EM algorithm was developed, capable of learning models of objects and object templates from snapshots of non-stationary objects, extracted from occupancy grid maps acquired at different points in time. Systematic experiments using a physical robot illustrate that our approach works well in practice, and that it outperforms a previously developed non-hierarchical algorithm for learning models of non-stationary objects.

Our approach possesses several limitations that warrant future research. For identifying non-stationary objects, our present segmentation approach mandates that objects do not move during robotic mapping, and that they are spaced far enough apart from each other (e.g., 5 cm). Beyond that, our approach currently does not learn attributes of objects other than shape, such as persistence, relations between multiple objects, and non-rigid object structures. Finally, exploring different generative models involving more complex transformations (e.g., scaling of templates) constitutes another worthwhile research direction.

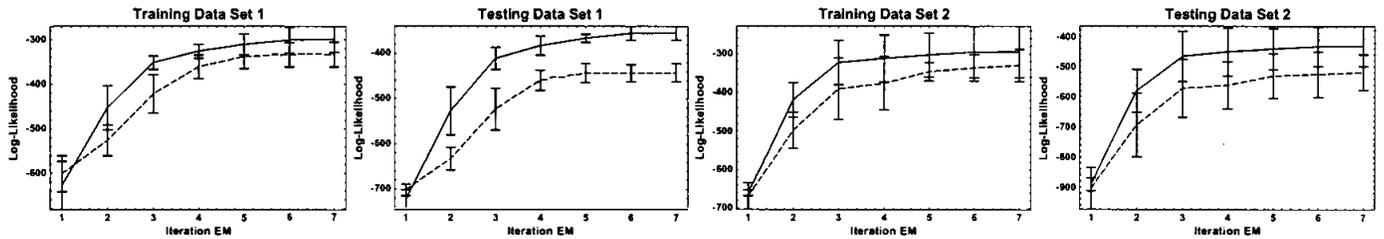

Figure 7: Log-likelihood of the training and testing (leave-one-out) data from both real-world data sets. The dashed line is the result of the shallow, non-hierarchical approach, which performs significantly worse than the hierarchical approach (solid line).

Nevertheless, we believe that this work is unique in its ability to learn hierarchical object models in mobile robotics. We believe that the framework of hierarchical models can be applied to a broader range of mapping problems in robotics, and we conjecture that capturing the object nature of robot environments will ultimately lead to much superior perception algorithms in mobile robotics, along with more appropriate symbolic descriptions of physical environments.

### Acknowledgements.

The work of Dragomir Anguelov and Daphne Koller was supported the Office of Naval Research, Young Investigator (PECASE) grant N00014-99-1-0464. Sebastian Thrun's work was supported by DARPA's MARS Program (Contract number N66001-01-C-6018) and the National Science Foundation (CAREER grant number IIS-9876136 and regular grant number IIS-9877033). This work was done while Sebastian Thrun was visiting Stanford University.